# Explainable AI for Ship Collision Avoidance:
## Decoding Decision-Making Processes and Behavioral Intentions

**Hitoshi Yoshioka ***   **Hirotada Hashimoto ***

*Division of Marine System Engineering, Graduate School of Engineering,*

*Osaka Metropolitan University*

*su23152i@st.omu.ac.jp

**hashimoto.marine@omu.ac.jp

**Abstract**

Most ship collision accidents are attributed to human errors. As long as humans operate ships, preventing these accidents remains a formidable challenge. Autonomous navigation technology is heralded as a potential solution to mitigate human error-related collisions. Recent advancements have enabled the application of deep reinforcement learning (DRL) in developing autonomous navigation artificial intelligence (AI). However, the decision-making process of AI is not transparent, and its potential for misjudgment could lead to severe accidents. Consequently, the explainability of DRL-based algorithms emerges as a critical hurdle in deploying autonomous collision avoidance systems that utilize AI. This study developed an explainable AI for ship collision avoidance. Initially, a critic network composed of sub-task critic networks was proposed to individually evaluate each sub-task in collision avoidance to clarify the AI decision-making processes involved. Additionally, an attempt was made to discern behavioral intentions through a Q-value analysis and an Attention mechanism. The former focused on interpreting intentions by examining the increment of the Q-value resulting from AI actions, while the latter incorporated the significance of other ships in the decision-making process for collision avoidance into the learning objective. AI's behavioral intentions in collision avoidance were visualized by combining the perceived collision danger with the degree of attention to other ships. The proposed method was evaluated through a numerical experiment. The developed AI was confirmed to be able to safely avoid collisions under various congestion levels, and AI's decision-making process was rendered comprehensible to humans. The proposed method not only facilitates the understanding of DRL-based controllers/systems in the ship collision avoidance task but also extends to any task comprising sub-tasks.

# 1. Introduction

Most ship collision accidents are attributed to human factors. Preventing such incidents remains challenging as long as humans operate ships. Maritime transportation safety is crucial for the global economy, given its significant role in international trade. Consequently, autonomous navigation technology is considered a potential solution to mitigate human error-related ship collisions. Essential tasks for realizing autonomous ship navigation include sailing efficiently to a set waypoint and safely avoiding collisions with other vessels. While the technology for sailing to a waypoint has been effectively developed through autopilot systems and can be implemented using classical controls such as PID control, collision avoidance requires more sophisticated decision-making. Developing a reliable model that mimics the advanced and flexible decision-making of human experts onboard is challenging. Collision avoidance decision-making involves assessing the risk of collision with other ships and choosing an appropriate course and/or speed. To date, rule-based collision avoidance algorithms using fuzzy logic[1] and the closest point of approach (CPA)[2] have been proposed. However, a standard algorithm for sophisticated decision-making applicable in various situations has not yet been established, as experts often make different judgments in congested scenarios. Therefore, deep reinforcement learning (DRL)-based collision avoidance has been researched recently[3–5], aiming to achieve decision-making akin to that of human experts. A DRL-based collision avoidance model autonomously acquires a decision-making capability, eliminating the need to explicitly describe it. However, the inability to explain the underlying decision-making process and behavioral intentions of AI presents a "black box" problem, leading to a lack of trust among seafarers. As such, autonomous collision avoidance AI cannot be used in actual ship navigation until this issue is resolved.

Explainable AI (XAI) has garnered significant attention recently, with various methods proposed. There are two primary types of XAI methods. One involves explaining the decision-making process of AI by analyzing changes in output when parts of the input are modified or missing[6–8]. However, these methods require extensive analyses with numerous input changes, rendering them excessively costly to perform in real-time. The other method attempts to identify the decision-making process by analyzing propagation in a neural network[9–13], although interpreting these results remains challenging for humans. Thus, a strong demand exists for methods that can explain the decision-making process of AI.

Given this background, this study developed an explainable AI for ship collision avoidance, focusing particularly on understanding the decision-making process and behavioral intentions. The decision-making process refers to how much collision danger the AI perceives from other ships, while behavioral intention indicates which ships the AI prioritizes to avoid. The explainable collision avoidance AI was developed using a deep deterministic policy gradient (DDPG)[14], suitable for continuous state and action spaces. An actor network and a critic network structure were proposed to clarify the AI's decision-making process. Behavioral intentions were analyzed using Q-value analysis and an attention mechanism, allowing for the visualization of AI's decision-making in collision avoidance as the product of perceived collision danger and attention to other ships. The effectiveness of the proposed method was confirmed through a numerical experiment.

## 2. Deep Reinforcement Learning
### 2.1. Reinforcement Learning

Reinforcement learning (RL) involves an agent interacting with an environment. During the learning process, the agent observes the current state of the environment, denoted as $s_t$. Based on this observation, the agent takes an action according to its policy, $\pi$, which results in the environment transitioning to the next state, $s_{t+1}$, and providing a reward, $r_{t+1}$, to the agent. The agent subsequently updates its policy with the goal of maximizing the cumulative future reward, using the reward received. The cumulative reward $R_t$ is calculated using Eq. (1), where $\gamma$ represents the discount rate, indicating the diminishing value of future rewards. This factor accounts for the uncertainty and decreasing significance of rewards received in later stages.

$$R_t = r_{t+1} + \gamma r_{t+1} + \gamma^2 r_{t+2} \ldots \tag{1}$$

Through extensive trial and error, the agent learns to choose actions that maximize this cumulative reward, ultimately determining the optimal action strategy.

### 2.2. Deep Deterministic Policy Gradient

In this study, the DDPG[14] algorithm, a form of DRL, is employed to develop an AI for collision avoidance. DDPG utilizes two neural networks: an actor network and a critic network. The critic network, denoted as $Q$, is responsible for learning the action value (Q-value) based on a given state and the action taken. Concurrently, the actor network, denoted as $\pi$, predicts the optimal action from the current state. During the learning process, the parameters of the critic network, $\theta_q$, are updated by minimizing the discrepancy between the predicted and the actual rewards received. The objective function used to train the critic network is outlined in Eq. (2). Additionally, DDPG employs target networks, $Q'$ and $\pi'$, which are similar to the structure of the critic and actor networks, respectively. The parameters of these target networks, $\theta'_q$ and $\theta'_\pi$, are periodically updated to stabilize the learning process, mitigating the risk of divergent or oscillating learning outcomes.

$$\mathcal{L}_{critic} = -\mathbb{E}\left[\left(r_{t+1} + \gamma Q(s_t, a_t|\theta_q) - Q'(s_{t+1}, \pi'(s_{t+1}|\theta'_\pi)|\theta'_q)\right)^2\right] \qquad (2)$$

The parameters of the actor network, $\theta_p$, are updated to maximize the Q-value as determined by the critic network. Consequently, the objective function for training the actor is expressed in Eq. (3). Since the parameters of the actor network are updated by minimizing the objective function, a minus sign is added to the mean Q-value.

$$\mathcal{L}_{actor} = -\mathbb{E}[Q(s_t, \pi(s_t|\theta_\pi))] \qquad (3)$$

Parameters of target networks are updated using Eq. (4), where $\alpha$ is the learning rate.

$$\begin{cases} \theta'_\pi \leftarrow \theta'_\pi + \alpha(\theta_\pi - \theta'_\pi) \\ \theta'_q \leftarrow \theta'_q + \alpha(\theta_q - \theta'_q) \end{cases} \qquad (4)$$

Additionally, a regularization term and a smoothing term[15] are incorporated into the objective function. A term that minimizes the scale of action is introduced to prompt the developed AI to select $a = 0$ when no differences in Q-value are apparent. Ultimately, the objective function for training the actor network is defined in Eq. (5), where $\hat{s}_t$, $\lambda_{reg}$, $\lambda_s$, and $\lambda_a$ represent the state with added random noise, a gain for regularization, a gain for smoothing, and a gain for minimizing the scale of action, respectively.

$$\mathcal{L}_{actor} = -\mathbb{E}[Q(s_t, \pi(s_t))] + \lambda_{reg} \sum \theta_\pi$$
$$+\lambda_s \sqrt{\mathbb{E}\left[(\pi(s_t) - \pi(\hat{s}_t))^2 + (\pi(s_t) - \pi(s_{t+1}))^2\right]} + \lambda_a \mathbb{E}[|\pi(s_t)|] \qquad (5)$$

## 3. Learning Environment
### 3.1 Ship Maneuvering Model
A coordinate system for ship motion is shown in Fig. 1.

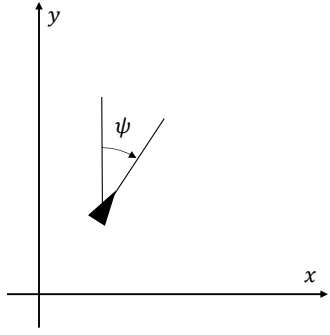

Figure 1 Coordinate system of ship motion

The motion of the ships is calculated using Nomoto's KT-model[16]. This motion is described by Eq. (6), where $r$, $\delta$, $K$, and $T$ represent the rate of turn, rudder angle, gain, and time constant, respectively. The position, $(x, y)$, and the heading angle, $\psi$, are updated according to Eq. (7).

$$T\dot{r} + r = K\delta \tag{6}$$

$$\begin{cases} \dot{x} = U\sin(\psi) \\ \dot{y} = U\cos(\psi) \\ \dot{\psi} = r \end{cases} \tag{7}$$

The action space of the agent is defined by the rudder angle, with a range from -5° to 5°.

### 3.2 Scenario Setting

The learning task involves navigating to a specified waypoint without colliding with other ships. The initial position, $(x_o, y_o)$, and course, $\psi_o$, are $(0[\text{km}], 0[\text{km}])$ and 0°, respectively. A waypoint is set at $(0\ [\text{km}], 46.3\ [\text{km}])$. Other ships are initialized randomly. The initial state is determined through several processes. Initially, the speed, $s_t$ [km/s], is randomly assigned within the range of 0.0041 to 0.0062 km/s. Next, the position of a collision with another ship, assuming both continue on a straight course, is calculated. The course over ground, $\psi_{col}$, is set to range from -150° to 150°. The time until collision, $t_{col}$, is randomly determined as 600–1800 s. Subsequently, the probable position of collision occurrence is calculated, as shown in Eq. (8). Random noises for position, $(dx, dy)$, and course, $d\psi$, are subsequently added. The noise ranges for position and course are −1 km to 1 km and −30° to 30°, respectively. The initial position of the other ship is determined as shown in Eq. (9). The initial state of this other ship is illustrated in Fig. 2.

$$\begin{cases} x_{col} = x_o + t_{col}(s_o \sin(\psi_o) - s_t \sin(\psi_{col})) \\ y_{col} = y_o + t_{col}(s_o \cos(\psi_o) - s_t \cos(\psi_{col})) \end{cases} \tag{8}$$

$$\begin{cases} x_o = x_{col} + dx \\ y_o = y_{col} + dy \\ \psi_o = \psi_{col} + d\psi \end{cases} \tag{9}$$

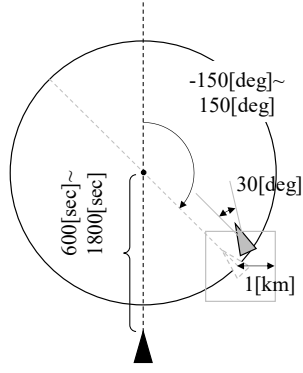

Figure 2 Initial state of other ships

The number of other ships is randomly determined, ranging from 0 to 6 ships. The time step for the simulation is set at 5 s, and the agent takes action at every step. Each episode consists of 240 steps.

### 3.3 Observation Setting

The observation includes information regarding the given waypoint and other ships. Specifically, the observation related to the waypoint, $s_t^{wp}$, is described in Eq. (10), where $(dx_{wp} \text{ [km]}, dy_{wp} \text{[km]})$ and $d\theta_{wp}$ [°] represent the relative position of the waypoint and the relative bearing angle toward it, respectively.

$$s_t^{wp} := [dx_{wp}, dy_{wp}, d\theta_{wp}] \tag{10}$$

The observation concerning the $i$-th other ship, $s_t^{oth_i}$, is expressed by Eq. (11), where $(dx^{oth_i} \text{ [km]}, dy^{oth_i} \text{ [km]})$ and $d\theta^{oth_i}$ are the relative distance and relative bearing angle, respectively. $(relv_x^{oth_i} \text{ [km/s]}, relv_y^{oth_i} \text{ [km/s]})$ is the relative speed. $D^{oth_i}$ [km] represents the distance from the own ship. Additionally, the observation includes information on a critical index for evaluating the danger of collision: the closest point of approach (CPA). The CPA represents the estimated closest point of encounter when both the own and other ship continue sailing straight. DCPA is the distance between own and another ship, and TCPA is the time taken by another ship to reach the CPA. DCPA and TCPA are commonly used by seafarers for making decisions regarding collision avoidance. The definition of CPA is illustrated in Fig.3, where OS and TS refer to own ship and another ship, respectively. The observation regarding the other ship included the relative position of CPA, $(dx_{CPA}^{oth_i} [km], dy_{CPA}^{oth_i}[km])$, DCPA, $DCPA^{oth_i}$ [km], and TCPA, $TCPA^{oth_i}$ [s].

$$s_t^{oth_i} := \begin{bmatrix} dx^{oth_i}, dy^{oth_i}, d\theta^{oth_i}, relv_x^{oth_i}, relv_y^{oth_i}, \\ D^{oth_i}, dx_{CPA}^{oth_i}, dy_{CPA}^{oth_i}, DCPA^{oth_i}, TCPA^{oth_i} \end{bmatrix} \quad (11)$$

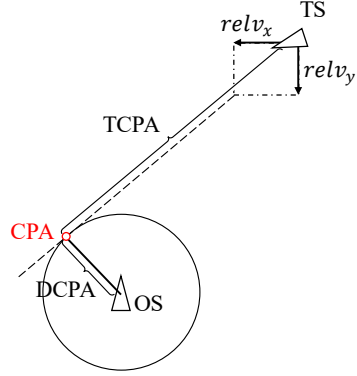

Figure 3 Definition of the closest point for approach

Finally, the observation data are obtained using Eq. (12), where $n$ is the number of other ships.

$$s_t := [s_t^{wp}, s_t^{oth_0}, \dots, s_t^{oth_n}] \quad (12)$$

The states of current and past four steps are provided as input to the actor and critic network.

### 3.3 Reward Setting

Rewards are assigned for successful collision avoidance and for navigating toward the specified waypoint. The reward for collision avoidance is determined based on the level of collision danger, which is evaluated using a ship domain that defines an exclusive area around the own ship. The level of danger, $cr^{oth_i}$, is calculated using Eq. (13), where $(ax_x, ax_y)$ represents the lengths of the x-axis and y-axis of the ship domain, respectively. $(ax_x, ax_y)$ is further explained in Eq. (14) and (15) using the ship's overall length, $loa = 0.34 \, [km]$. The shape of this domain is illustrated in Fig. 4.

$$cr_t^{oth_i} = 1 - \sqrt{(dx_t^{oth_i}/ax_x)^2 + (dy^{oth_i}/ax_y)^2} \quad (13)$$

$$ax_x = \begin{cases} 3.2lpp, & dx_t^{oth_i} \geq 0 \\ 1.6lpp, & dx_t^{oth_i} < 0 \end{cases} \quad (14)$$

$$ax_y = \begin{cases} 6.4lpp, & dy^{oth_i} \geq 0 \\ 1.6lpp, & dy^{oth_i} < 0 \end{cases} \quad (15)$$

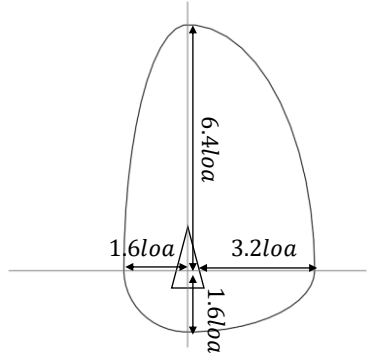

Figure 4 Shape of the exclusive area

The reward associated with the collision risk involving the $i$-th other ship, $r_t^{oth_i}$, is calculated using Eq. (16). To promote actions that reduce the danger of collision, a higher penalty is applied as the danger of collision increases.

$$r_t^{oth_i} = \begin{cases} 0.0, & cr_t^{oth_i} = 0 \\ -0.5 * cr_t^{oth_i} - 0.1, & cr_t^{oth_i} \neq 0 \ \& \ cr_t^{oth_i} \leq cr_{t-1}^{oth_i} \\ -1.0 * cr_t^{oth_i} - 0.5, & cr_t^{oth_i} \neq 0 \ \& \ cr_t^{oth_i} > cr_{t-1}^{oth_i} \end{cases} \quad (16)$$

The reward for navigating toward the given waypoint is calculated using Eq. (17), based on the relative bearing angle toward the waypoint, denoted as $\theta_t^{wp}$.

$$r_{wp} = 0.20 * \left(0.9 \exp\left(-\left(d\theta_t^{wp}/60\right)^2/2\right) + 0.1\left(1 - |d\theta_t^{wp}|/180\right)\right) \quad (17)$$

To incentivize actions that reduce the discrepancy between the own ship's course and the direction toward the given waypoint, a reward based on the reduction of $\theta_t^{wp}$, denoted as $r_{\Delta wp}$, is awarded when the sum of danger levels is zero. This reward $r_{\Delta wp}$ is defined in Eq. (18). The function $clip(\cdot, a, b)$ modifies values, such that any value less than $a$ is converted to $a$, and any value greater than $b$ is converted to $b$.

$$r_{\Delta wp} = 0.050 * clip\left(\frac{(|d\theta_{t-1}^{wp}| - |d\theta_t^{wp}|)}{\min(10, |d\theta_{t-1}^{wp}|)}, 0, 1\right) \tag{18}$$

## 4. Explanation of Decision-Making Process
### 4.1 Critic Network Consisted of Sub-task Critics

An autonomous navigation task involves heading to a specified waypoint while avoiding collisions with other ships. Traditional collision avoidance AI systems learn these tasks collectively by processing a combined reward for both navigating to the waypoint and avoiding collisions. Actions to prevent collisions are initiated when the AI deems another ship as a danger, rendering it crucial to identify which ships are considered dangerous for understanding the underlying decision-making process. However, existing AIs cannot compute the reward components for each task within the predicted Q-value.

To address this limitation, a novel approach involving a critic network capable of outputting separate Q-values for each sub-task is proposed and has been validated through numerical experiments. This network is referred to as "sub-task critics" (STC) throughout the paper. The STC utilizes distinct input data for each sub-task, ensuring that the components of the Q-value are dependent solely on the relevant input data. This design allows the STC to learn specific aspects of Q-values related to each sub-task independently.

The sub-tasks in this autonomous navigation scenario include sailing to the designated waypoint and avoiding collisions. The architecture of the critic network, $Q$, is defined in Eq. (19) and includes separate critic networks for each sub-task: for navigating to the waypoint, $Q_{wp}(s^{wp}, a|\theta_{wp})$, and for collision avoidance, $Q_{ca}(s^{oth}, a|\theta_{ca})$, where $\theta_{wp}$ and $\theta_{ca}$ are the trainable parameters of the STC. The objective function for the STC remains the same as in the original DDPG method. Since the STC learns the components of the Q-value based on specific input data for each sub-task, individual rewards need not be designed for each task, simplifying the training process.

$$Q(s, a|\theta_q^{wp}, \theta_q^{ca}) = Q_{wp}(s^{wp}, a|\theta_q^{wp}) + \sum_i Q_{ca}(s^{oth_i}, a|\theta_q^{ca}) \tag{19}$$

The calculation flow within the critic network is shown in Fig. 5.

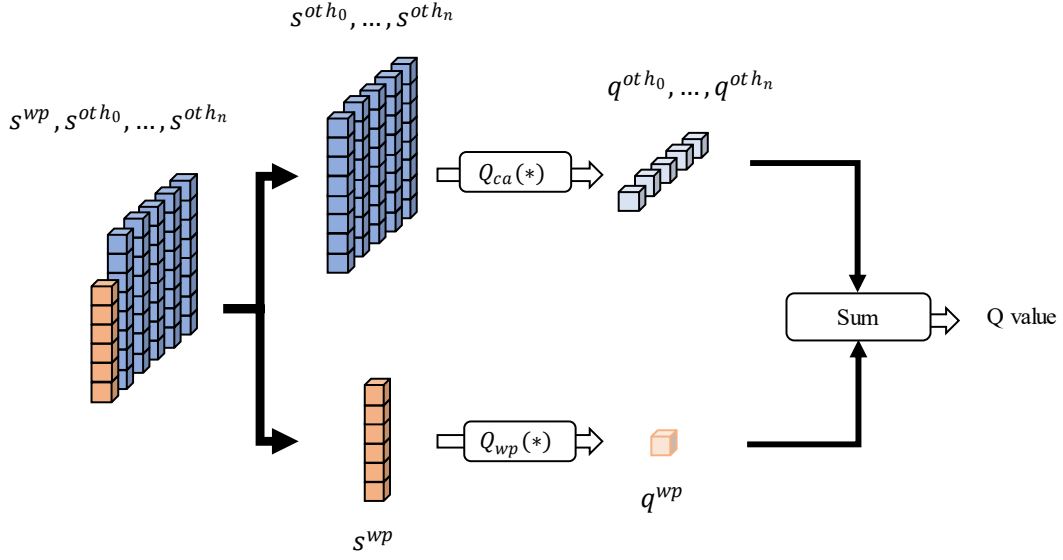

Figure 5 Overview of calculation flow in the proposed critic network

**4.2 Actor Network for Variable Length of Observation**

In this study, the observation space, as defined by Eq. (11), varies in size depending on the number of other ships surrounding the own ship. Typically, differences in input size are managed by padding the data to achieve uniformity or by designing inputs that mimic image data. However, padding introduces unnecessary data, which can potentially influence the output, and the use of image-like data may lead to a loss of accuracy depending on the grid size, in addition to imposing a limit on the maximum number of ships that can be accommodated.

To address these challenges, a structure for the actor network capable of handling variable-sized input data is proposed. This network consists of a two-step calculation process. The first step involves an encoding phase where observations for each task are converted into uniformly sized features using task-specific neural networks. Subsequently, an integrated feature set is computed by combining these encoded features. The final step involves predicting the optimal action based on the integrated features.

In this framework, neural networks encoding $s^{wp}$ and $s^{oth}$ are employed for encoding the observation state. The integrated features, represented as $F$, are expressed by Eq. (20), where $E_{wp}(s^{wp}|\theta_e^{wp})$ and $E_{oth}(s^{oth}|\theta_e^{oth})$ are neural networks encoding $s^{wp}$ and $s^{oth}$. A softsign function is applied to scale the total of the encoded features within the range of -1 to 1. $\theta_e^{wp}$ and $\theta_e^{oth}$ are the parameters of this function.

$$F(s) = \text{softsign}\left(E_{wp}(s^{wp}|\theta_e^{wp}) + \sum_i E_{oth}(s^{oth_i}|\theta_e^{oth})\right) \quad (20)$$

A neural network that predicts an optimal action based on integrated features is denoted as $A(F|\theta_a)$, where $\theta_a$ represents the parameters of the network. The function of the actor network, $\pi$, is expressed by Eq. (21).

$$\pi(s|\theta_e^{wp}, \theta_e^{oth}, \theta_a) = A\left(E_{wp}(s^{wp}|\theta_e^{wp}) + \sum_i E_{oth}(s^{oth_i}|\theta_e^{oth}) | \theta_a\right) \quad (21)$$

The workflow within the actor network is depicted in Fig. 6. Utilizing the structure of the actor network, expressed by Eq. (21), the network can adapt its predictions even when the number of ships changes.

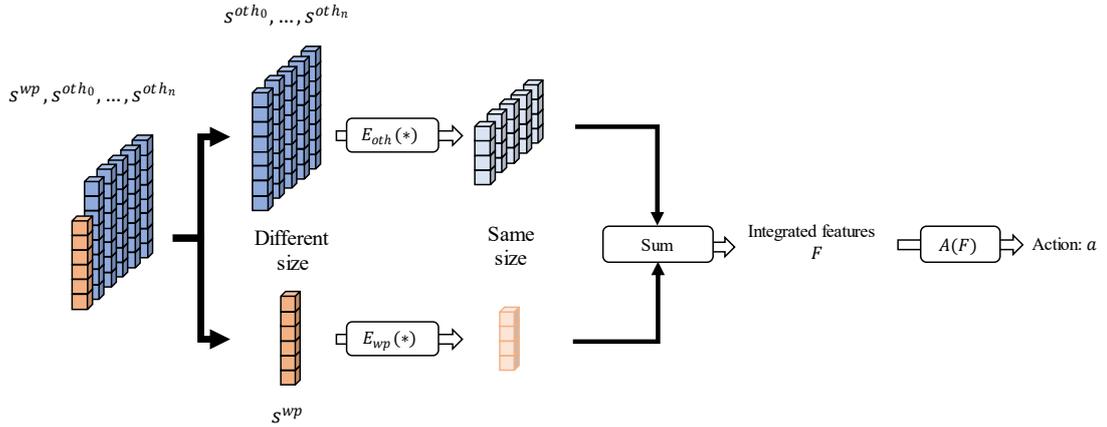

Figure 6 Overview of calculation flow in the proposed actor network

**4.2 Results and Discussion**

The proposed method was validated through a numerical experiment. The trajectories from the numerical experiment results are depicted in Fig. 7, while the relative trajectories are illustrated in Fig. 8. Filled markers represent other ships, and unfilled markers indicate the motion of own ships.

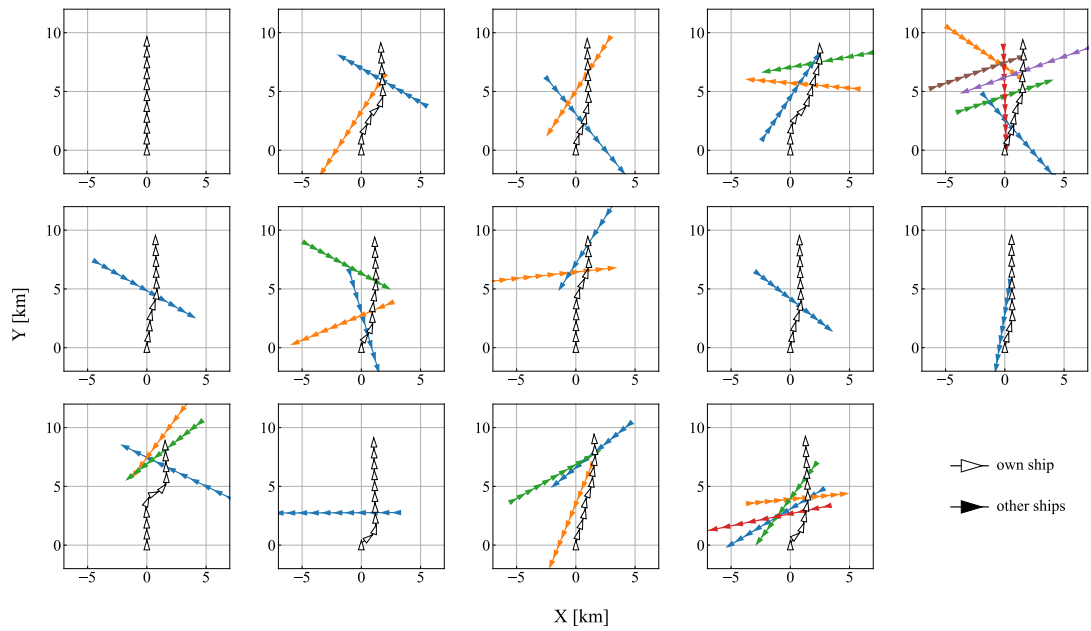

Figure 7 Trajectories of the numerical experiment results

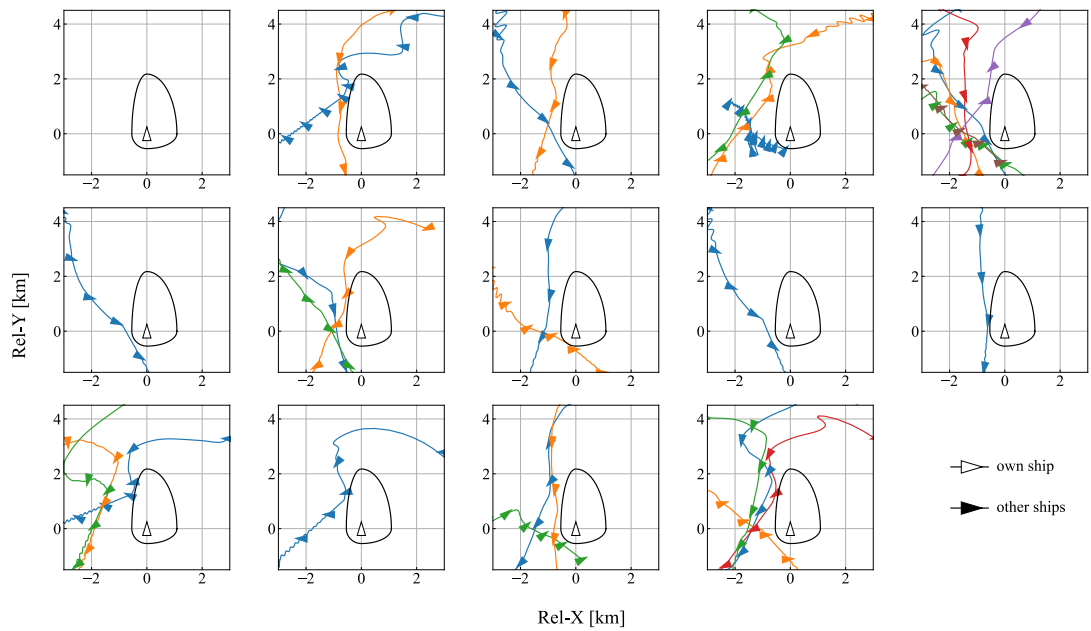

Figure 8 Relative trajectories of the numerical experiment results

According to the results, the developed AI successfully prevents the invasion of its exclusive area by other ships, demonstrating that the AI has learned safe collision avoidance maneuvers.

The relative trajectories, colored by evaluation values assigned by the critic, are shown in Fig. 9. The danger of collision increases as the evaluation value decreases. The results indicate that when another ship is near the domain of the own ship and is closing in, it is judged as a danger. Fig. 9 visually represents which ships are deemed dangerous using the proposed critic network.

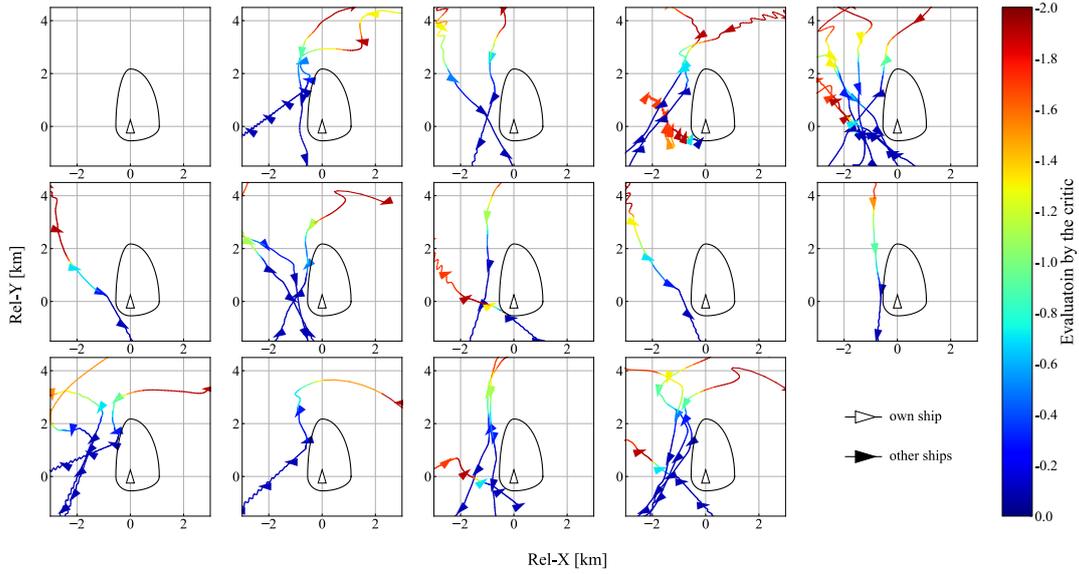

Figure 9 Relative trajectories of the numerical experiments colored based on evaluation by the critic

## 5. Interpretation of Behavioral Intention
### 5.1 Behavioral Intention

In the previous section, a method was proposed to explain the evaluation value for each task, demonstrating how the collision avoidance AI assesses the danger level of each ship. In this section, the behavioral intentions are visualized. In this study, behavioral intention refers to the actions the AI aims to take, such as the choice of ships to avoid or the preference to perform collision avoidance maneuvers or proceed to a given waypoint.

### 5.2 Interpretation Using Q-value

The actor network trains to select an optimal action by minimizing the objective function, as indicated in Eq. (4). Under conditions of no imminent danger or specific waypoint, the AI opts to sail straight ($\delta = 0°$), as dictated by the fourth term of the objective function. However, if another action is chosen, AI has a specific objective. According to the DRL framework, this objective is to maximize the Q-value; therefore, the aim of choosing an action is to increase the Q-value. This objective can be interpreted based on whether the Q-value increases or decreases. Specifically, a task that results in an increase in Q-value can be considered a goal the AI seeks to achieve.

The difference in Q-value between the taken action and sailing straight, $a = 0$, is expressed in Eq. (22) and is referred to as the priorities of sub-tasks.

$$\begin{cases} p_{wp} = Q_{wp}(s^{wp}, a) - Q_{wp}(s^{wp}, 0) \\ p_{oth} = Q_{wp}(s^{oth}, a) - Q_{wp}(s^{oth}, 0) \end{cases} \quad (22)$$

The trajectories and the time series of properties for the scenario depicted in the fourth figure of Fig. 7 are shown in Fig. 10. Additionally, the encounter situation, actions taken, and priorities at 600 s and 750 s are illustrated in Fig. 11. According to Fig. 11, at 600 s, the own ship turns right to avoid collisions, and the priorities for collision avoidance increase. Subsequently, at 750 s, as the property for sailing toward the given waypoint increases and the priorities for collision avoidance decrease, AI's intention can be understood as the own ship turning left to sail toward the waypoint, despite an increase in the potential danger of collision.

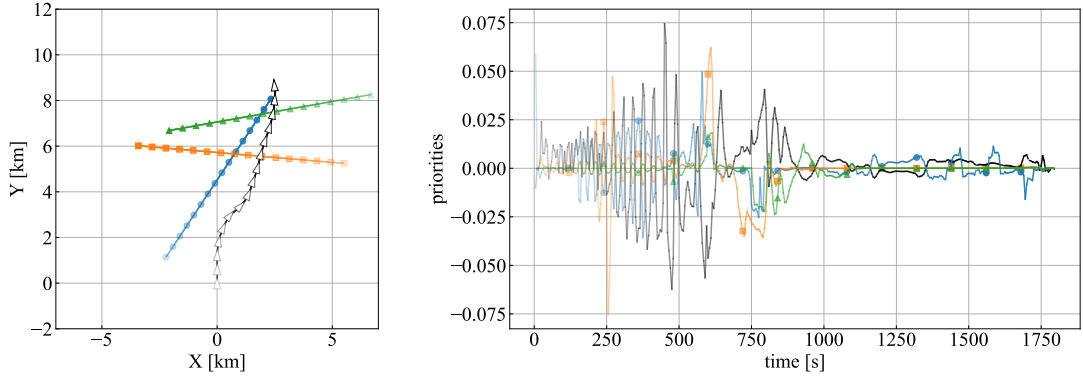

Figure 10 Trajectories and timeseries of priorities in the fourth scenario

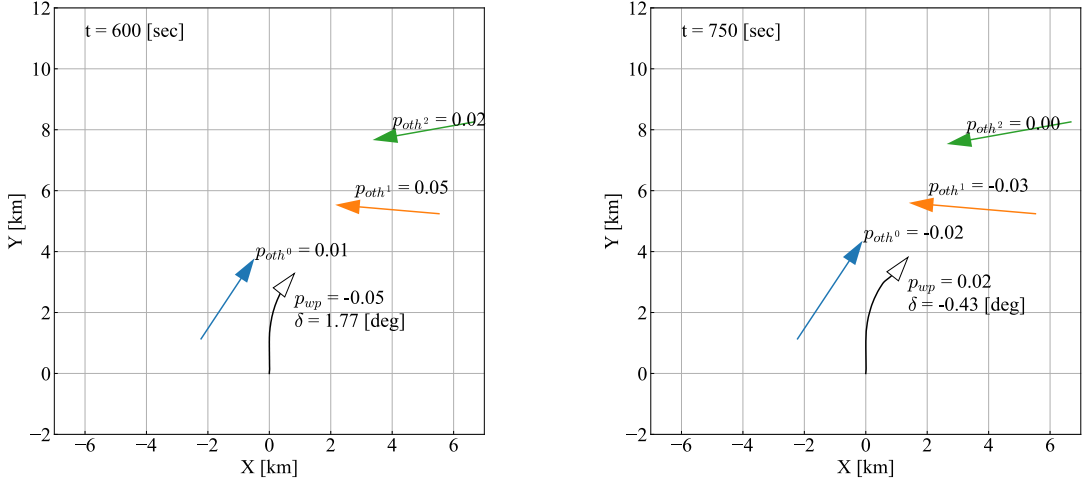

Figure 11 Encounter situation and priorities at 600 s and 750 s

## 5.3 Interpretation Using Attention Mechanism

In the previous section, the interpretation of behavioral intention using Q-values was conducted. However, these interpretations are not guaranteed to align precisely with the AI's intentions. Therefore, incorporating the AI's intention into the objective of learning becomes essential. In deep learning, the appropriate focus of AI is learned using an Attention mechanism. This mechanism has been integrated into the actor network to enable the learning of attention values for other ships. The attention value, $\alpha^{oth}$, is calculated using a network called Attention, as expressed by Eq. (23). The $Attention$ mechanism receives as input a set of own ship's state and another ship's state, denoted as $S_\alpha^{oth_i} \coloneqq \left[[s^{oth_i}, s^{oth_0}], [s^{oth_i}, s^{oth_1}], \ldots, [s^{oth_i}, s^{oth_n}]\right]$. Subsequently, it processes the input to produce the aggregated results of applying $Attention$ across all $S_\alpha^{oth_j}$. Finally, the attention values are normalized to sum to 1 using a softmax function to calculate the total attention value.

$$[\alpha^{oth_0}, \ldots, \alpha^{oth_n}]$$
$$= softmax\left(\left[\sum_j Attention(s^{oth_i}, s^{oth_j}|\theta_\alpha), \ldots, \sum_j Attention(s^{oth_n}, s^{oth_j}|\theta_\alpha)\right]\right) \quad (23)$$

The attention values are utilized to integrate the encoded features of other ships. In the actor network structure depicted in Fig. 6, the integration of encoded features from other ships is accomplished by calculating their sum. Conversely, in the structure that incorporates the Attention mechanism, the encoded features of other ships are integrated by calculating a weighted average using the attention values. An overview of this calculation process is shown in Fig. 12.

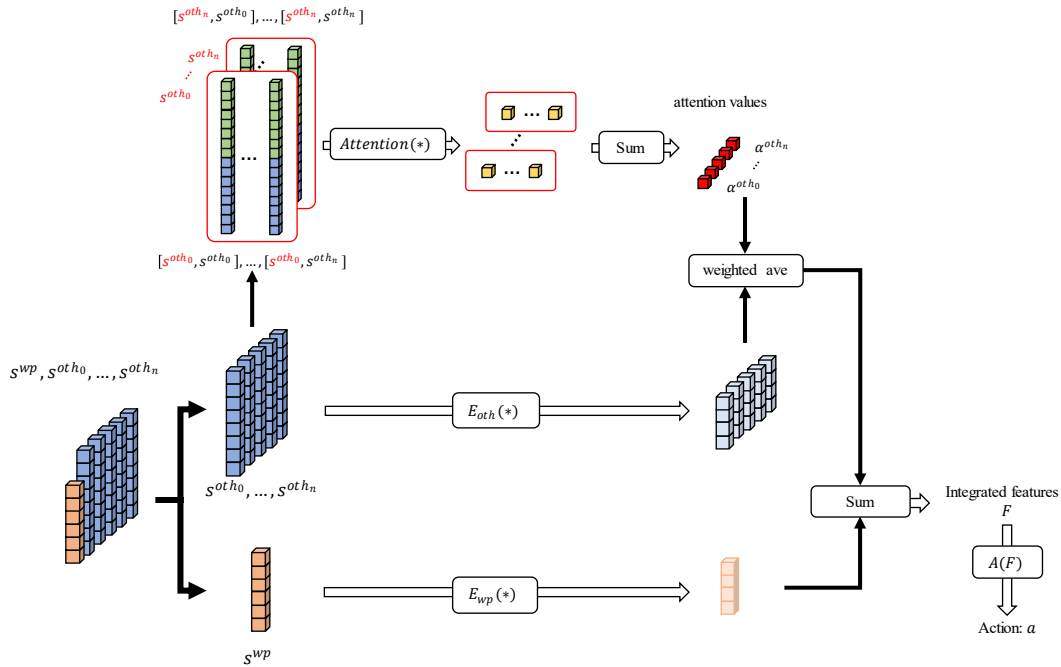

Figure 12. Overview of calculations in the actor network underlying the attention mechanism

The attention values of the well-trained actor network, in the scenario depicted in Fig. 10 and 11, are illustrated in Fig. 13. The trajectories of other ships are color-coded based on their attention values. Ships with higher attention values exhibit a greater influence on decision-making processes. According to the results, initially, own ship focused on the ship to the left. Subsequently, attention shifted to a ship crossing from the right side, which own ship successfully avoided. After passing these ships, the overtaking ship from the left side garnered attention, and own ship passed it. These results visually demonstrate which other ships significantly impact own ship's decision-making for collision avoidance.

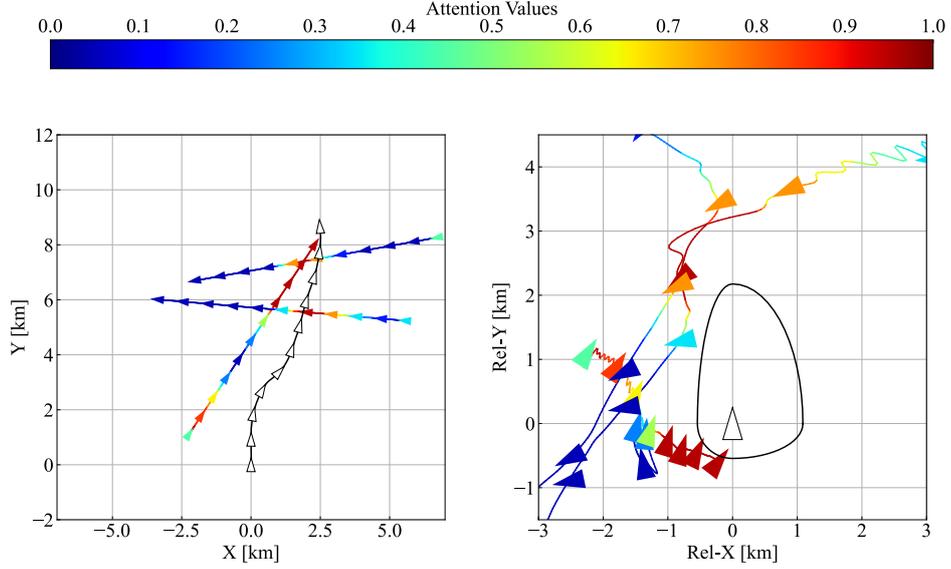

Figure 13 Absolute trajectories and relative trajectories colored based on attention values.

The use of the *Attention* mechanism enables the contributions of states concerning other ships to be quantified. However, the attention values only indicate the relative importance among the ships, highlighting which ships are being focused on at any given time. For instance, as shown in Fig. 13, the overtaking ship receives almost all the attention from the actor network once all collisions have been avoided. However, a ship with a low collision risk may not be relevant for decision-making related to collision avoidance if it continues to receive attention. Consequently, attention values alone are not synonymous with behavioral intentions.

Therefore, behavioral intentions are calculated by considering both the attention values and the evaluation values provided by the critic network. An index that quantifies the contribution of the i-th other ship to the behavioral intention, denoted as Intention, is derived using Eq. (24). This approach ensures a more comprehensive assessment of each ship's impact on the AI's decision-making process.

$$Intention^{oth_i} = Q_{ca}(s^{oth_i}) \times Attention(s^{oth_i}) \qquad (24)$$

Eq. (24) is applied to the scenario depicted in the fourth figure of Fig. 7. Both absolute and relative trajectories, colored according to the value of $Intention^{oth_i}$, are illustrated in Fig. 14. The ship displayed in the lower-left of the relative trajectory in Fig. 14 did not influence the decision-making process. In conclusion, the contributions of other ships to decision-making were quantitatively identified using the attention value and the evaluation value by the critic. The

Intention value aids in understanding the AI's thought process in making decisions for ship collision avoidance.

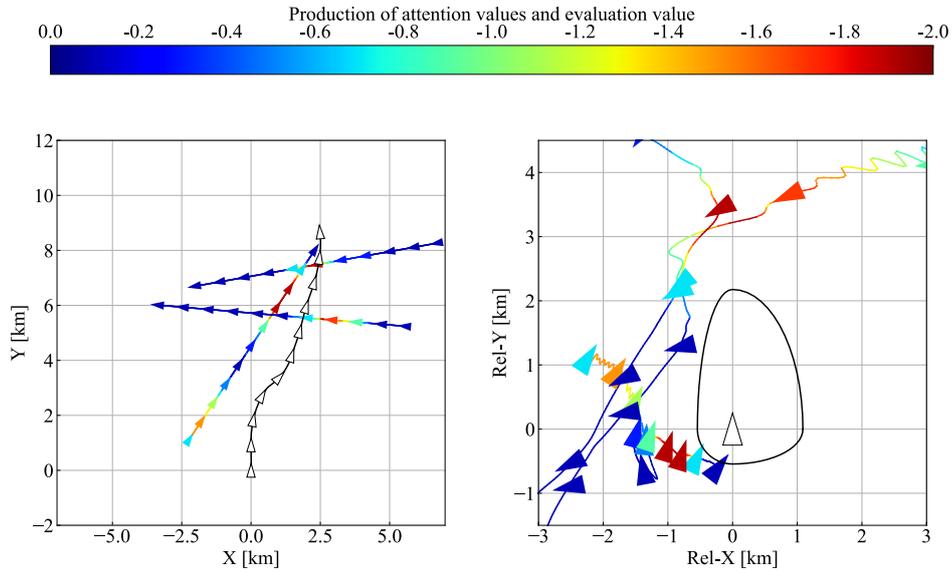

Figure 14 Absolute trajectories and relative trajectories colored based on intention values.

### 5.4 Time Variation of Intention

To elucidate AI's intention in ship collision avoidance, two methods were proposed: one utilizing the Q-value described in Section 5.2, and another employing an Attention mechanism outlined in Section 5.3. These methods were applied to the same encounter scenario to compare their effectiveness and differences. The time series of both methods, including the Q-values, attention values, and trajectories, are displayed in Fig. 15. The results show a similar trend at approximately 600 s where both $p_{wp}$ and Intention, indicated by orange squared markers, increase. However, their trends from 200 to 500 s differ. Further evaluation of these methods requires expert collaboration in the future. Additionally, an integrated model is anticipated as future work, which can more clearly explain AI's behavioral intentions.

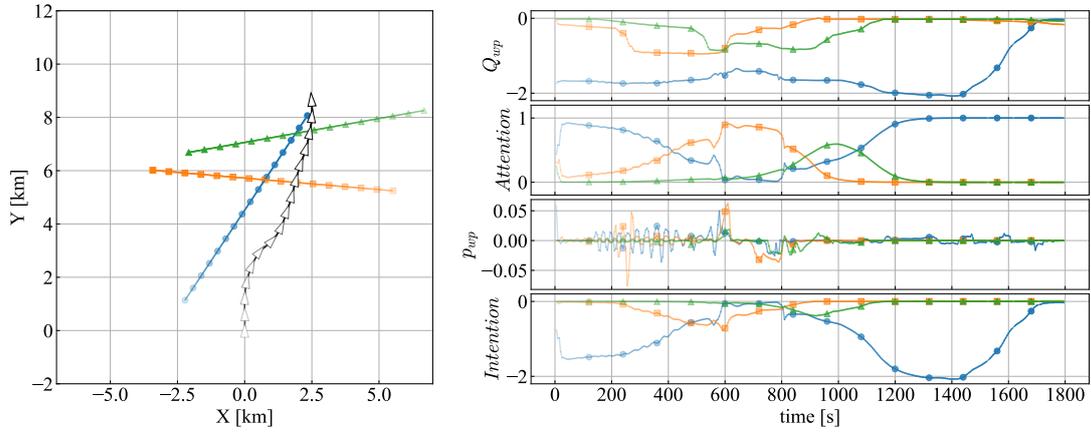

Figure 15 Time histories of indexes of intention.

## 6 Conclusions

The development of AI using DRL has been explored by numerous researchers for autonomous ship navigation. AI is anticipated to facilitate sophisticated and flexible decision-making comparable to that of human experts in various encounter scenarios. However, implementing AI in practical settings is challenging because AI-based systems are perceived as black boxes by seafarers, and any misjudgment by AI could result in severe accidents. To address this issue, this study proposes an explainable collision avoidance AI system.

The autonomous navigation task is subdivided into tasks such as sailing to a designated waypoint and avoiding collisions with other ships. A critic network composed of sub-critic networks, which evaluate the Q-value of each sub-task, was proposed to clarify the decision-making process of AI. Given that the size of the input state varies with the number of other ships, an actor network capable of handling variable-length input data was utilized. This network includes encoders and an actor, which standardize the data size of observations about other ships and waypoints into uniform feature sizes. These features are aggregated by summing them, and the actor network then computes actions based on these integrated features. The AI was trained using the critic and actor networks to learn collision avoidance maneuvers, and its performance was assessed through numerical experiments. The results allow the visualization of the danger of collision as evaluated by the AI, enabling humans to comprehend the perceived collision danger under varying circumstances.

Additionally, the study sought to render the behavioral intentions of AI understandable. The critic network is capable of calculating individual Q-values for the danger of collision with other ships, and the intention was interpreted by assessing the increase in Q-value from the base scenario of sailing straight (no action). While the action resulting in the largest increase in Q-value could be considered indicative of the AI's desired action, it does not necessarily reflect the actual intention. Therefore, the importance of each ship in the decision-making process for collision avoidance was incorporated into the learning objective. These importances were learned using an Attention mechanism integrated into the actor network. As a result, the contributions of other ships to the decision-making process were elucidated. The behavioral intention of the AI was interpreted as a combination of the attention value and the evaluation value (collision danger). Several examples of behavioral intentions were presented in a numerical experiment.

The proposed methods have made the decision-making process and behavioral intentions of the collision avoidance AI understandable. Notably, the computational cost associated with calculating the behavioral intention is minimal, allowing for real-time calculation and visualization. The developed explainable AI for collision avoidance is comprehensible to seafarers onboard and could significantly contribute to the future practical implementation of autonomous navigation AI systems.

**Acknowledgement**

This work was supported by JST SPRING Grant Number JPMJSP2139 and JSPS KAKENHI Grant Number 23H01627.

# Appendix
**Hyperparameters for learning**

Training was conducted using the actor and critic network described in the main text. The neural network structure of the $Q_{wp}(s^{wp}, a|\theta_q^{wp})$, $Q_{ca}(s^{oth_i}, a|\theta_q^{ca})$, $E(s^{wp}|\theta_e^{wp})$, $E(s^{oth}|\theta_e^{oth_i})$, and $A(F|\theta_a)$ are presented below. The leak rate for Leaky ReLU is 0.1, and other hyperparameters are shown in Table A.6.

Table A.1 Neural Network Structure of $Q_{wp}(s^{wp}, a|\theta_q^{wp})$

| | |
|---|---|
| Fully Connected | 256 |
| ReLU | - |
| Fully Connected | 256 |
| ReLU | - |
| Fully Connected | 128 |
| ReLU | - |
| Fully Connected | 1 |

Table A.2 Neural Network Structure of $Q_{ca}(s^{oth_i}, a|\theta_q^{ca})$

| | |
|---|---|
| Fully Connected | 256 |
| ReLU | - |
| Fully Connected | 256 |
| ReLU | - |
| Fully Connected | 128 |
| ReLU | - |
| Fully Connected | 1 |

Table A.3 Neural Network Structure of $E(s^{wp}|\theta_e^{wp})$

| | |
|---|---|
| Fully Connected | 256 |
| Leaky ReLU | - |
| Fully Connected | 256 |
| Leaky ReLU | - |
| Fully Connected | 128 |

Table A.4 Neural Network Structure of $E(s^{oth}|\theta_e^{oth_i})$

| | |
|---|---|
| Fully Connected | 256 |
| Leaky ReLU | - |
| Fully Connected | 256 |
| Leaky ReLU | - |
| Fully Connected | 128 |

Table A.5 Neural Network Structure of $A(F|\theta_a)$

| | |
|---|---|
| Fully Connected | 256 |
| Leaky ReLU | - |
| Fully Connected | 256 |
| Leaky ReLU | - |
| Fully Connected | 128 |
| Leaky ReLU | - |
| Fully Connected | 1 |
| softsign | - |

Table A.6 Hyperparameters for learning

| | |
|---|---|
| Learning steps | 10,000,000 |
| Train interval | 100 |
| Learning rate | 0.001 |
| Target model update rate | 0.001 |
| Batch size | 1024 |
| Optimizer | Adam |
| Discount ratio | 0.98 |
| $\lambda_{reg}$ | 0.001 |
| $\lambda_s$ | 0.005 |
| $\lambda_a$ | 0.0001 |